\documentclass[runningheads]{llncs}

\usepackage[T1]{fontenc}
\def\doi#1{\href{https://doi.org/\detokenize{#1}}{\url{https://doi.org/\detokenize{#1}}}}

\usepackage{multirow}
\usepackage{tabularx}
\usepackage{amssymb}
\usepackage{amsmath}
\usepackage{bm}
\usepackage{graphicx}
\usepackage{amsfonts}
\usepackage{booktabs}
\usepackage{array}
\usepackage{algpseudocode}
\usepackage{siunitx}
\usepackage{cite}
\usepackage[colorlinks = true,
   linkcolor = blue,
   urlcolor  = black,
   citecolor = black,
   anchorcolor = blue]{hyperref}
\usepackage{mathtools}
\usepackage[misc]{ifsym}
\usepackage{pifont}

\usepackage{gensymb}
\usepackage{mathpazo} 

\newcolumntype{C}[1]{>{\centering\let\newline\\\arraybackslash\hspace{0pt}}m{#1}}
\usepackage{setspace}

\usepackage{listings}
\usepackage[a4paper, top=3cm, bottom=3cm, left=2.5cm, right=2.5cm]{geometry} 

\begin{document}
\onehalfspacing

\title{SegHeD+: Segmentation of Heterogeneous Data for Multiple Sclerosis Lesions with Anatomical Constraints and Lesion-aware Augmentation}
\titlerunning{SegHeD+}
%
\author{Berke Doga Basaran\inst{1,2}\textsuperscript{(\Letter)} \and 
Paul M. Matthews\inst{3,4,5} \and \\
Wenjia Bai\inst{1,2,3}}
%

\authorrunning{B.D. Basaran et al.}
%
\institute{Department of Computing, Imperial College London, London, UK \and
Data Science Institute, Imperial College London, London, UK \and
Department of Brain Sciences, Imperial College London, London, UK \and 
UK Dementia Research Institute, Imperial College London, London, UK \and
Rosalind Franklin Institute, Didcot, UK \\
\email{\Letter \hspace{.1cm} bdb19@imperial.ac.uk} }


%
\maketitle
\begin{abstract} 
Assessing lesions and tracking their progression over time in brain magnetic resonance (MR) images is essential for diagnosing and monitoring multiple sclerosis (MS). Machine learning models have shown promise in automating the segmentation of MS lesions. However, training these models typically requires large, well-annotated datasets. Unfortunately, MS imaging datasets are often limited in size, spread across multiple hospital sites, and exhibit different formats (such as cross-sectional or longitudinal) and annotation styles. This data diversity presents a significant obstacle to developing a unified model for MS lesion segmentation. To address this issue, we introduce SegHeD+, a novel segmentation model that can handle multiple datasets and tasks, accommodating heterogeneous input data and performing segmentation for all lesions, new lesions, and vanishing lesions. We integrate domain knowledge about MS lesions by incorporating longitudinal, anatomical, and volumetric constraints into the segmentation model. Additionally, we perform lesion-level data augmentation to enlarge the training set and further improve segmentation performance. SegHeD+ is evaluated on five MS datasets and demonstrates superior performance in segmenting all, new, and vanishing lesions, surpassing several state-of-the-art methods in the field.
\end{abstract}

\section{Introduction} 
Multiple sclerosis (MS) is a central nervous system disorder characterized by inflammation and myelin loss~\cite{Lassmann2018}. The accurate identification and measurement of MS lesions in brain MR images is essential for clinical diagnosis and research. While many lesion segmentation methods have been proposed in recent years, such as those based on machine learning, they are often trained on well-curated datasets with a consistent data format~\cite{Carass2015, Commowick2018, Commowick2021}. 

Clinical studies on MS involve gathering imaging data from multiple sources, resulting in various image qualities due to magnetic resonance machine and hyperparameter differences. The diversity in acquired images makes developing robust and well-generalising models paramount for clinical adoption. However, MS datasets curated for lesion segmentation vary in formats for both data and annotations, for example cross-sectional or longitudinal image datasets. The inability of off-the-shelf models adapting to heterogeneous data formats reduce their capability of including more and diverse data, resulting in domain-shift problems when they are deployed on unseen data. There is a lack of specialized approaches designed to handle lesion segmentation when dealing with such diverse data and annotation formats. Efforts to bridge this gap are crucial for advancing the performance and applicability of lesion segmentation models in MS research and clinical practice.

Detailed segmentation of MS lesions is critical for monitoring disease progression, evaluating treatments, and guiding personalised care. While machine learning-based segmentation methods are improving with new datasets and model advancements, their applications are often clinically inadequate, being only capable of all lesion segmentation. Most publicly available datasets focus on single timepoint scans, despite the temporal nature of neuro-degenerative diseases and importance of tracking disease progression for clinical decisions. For example, vanishing lesions, which may delay diagnosis and reflect transient disease activity or repair, are frequently overlooked due to limited longitudinal imaging data~\cite{Eckert2018}. Moreover, MS subtypes show differing lesion dynamics; relapsing-remitting MS features more new lesion formation, while progressive MS exhibits higher lesion atrophy rates, closely linked to disability progression~\cite{Dwyer2018}. These gaps highlight the need for models which integrate temporal lesion dynamics and attention to lesion sub-types to enhance clinical applicability.

In this study, we introduce SegHeD+, a novel brain lesion segmentation model that learns from multiple datasets and for multiple tasks simultaneously, leveraging heterogeneous data for model training. SegHeD+ enables the incorporation of information from both single-timepoint (cross-sectional) and multiple-timepoint (longitudinal) images, as well as from annotations for various types of lesions, including all, new, and vanishing lesions~\cite{prineas199}. Our model integrates temporal and volumetric constraints, along with anatomical constraints, to improve the plausibility of the segmentated lesions. In model training, we incorporate a lesion-level data augmentation method, which allows the generation of synthetic lesion images to enlarge the training set. Experimental results demonstrate that SegHeD+ achieves competitive performance compared to current state-of-the-art (SOTA) approaches for MS lesion segmentation.

\subsection{Related works}
\subsubsection{Longitudinal lesion segmentation} 
Several advancements have been made in machine learning techniques for segmenting cross-sectional brain lesion images~\cite{kamnitsas2017, basaran2023}. However, longitudinal lesion segmentation, which capitalizes on temporal information within longitudinal imaging data, remains relatively underexplored. Elliott et al. utilises the difference map between two timepoints to identify new lesions at the second time point~\cite{Elliott2013}. Jain et al. also leverages the difference map and devises an expectation maximization framework to jointly segment lesions at both timepoints~\cite{Jain2016}. Denner et al. incorporates a displacement field to capture spatiotemporal changes between timepoints~\cite{Denner2021}. Basaran et al. employs an nnU-Net model~\cite{Isensee2021} in conjunction with lesion-aware data augmentation to identify new lesions at the second timepoint~\cite{Basaran2022}.

\subsubsection{Learning from heterogeneous data}
Publicly available datasets for brain lesion segmentation come in various formats and follow different annotation protocols. Some datasets are cross-sectional, consisting of scans from a single timepoint, while others are longitudinal, including scans from two or more timepoints. This diversity poses a significant challenge when developing a universal machine learning model that can accommodate different data formats. Wu et al. suggests learning from diverse data for both all-lesion and new-lesion segmentation by incorporating relation regularization into the prediction of all lesions~\cite{Wu2023}. Liu et al. develops a multi-organ segmentation model by training it on 13 different organ datasets and integrating CLIP-inspired label encoding techniques~\cite{Liu2023}. Shi et al. introduces a marginal loss and an exclusion loss to train a unified multi-organ segmentation model using multiple partially annotated datasets~\cite{Shi2021}. With the increasing amount of data and available resources, training a universal model from multiple heterogeneous datasets is becoming more and more popular whilst showcasing impressive results~\cite{Butoi_2023_ICCV, lin2023samus}.

\subsubsection{Enforcing anatomical plausibility}
The incorporation of anatomical plausibility within machine learning models enhances the reliability and trustworthiness of the trained models for clinical use. Anatomical constraints have been introduced into machine learning frameworks in a variety of techniques. Dalca et al. introduces a model incorporating a variational auto-encoder to acquire location-specific priors for segmenting brain structures~\cite{Dalca2018}. Strumia et al. restricts lesion segmentation to the white matter by leveraging a geometric brain model~\cite{Strumia2016}. Lastly, Hirsh et al. utilises a multi-prior network alongside tissue probability maps for segmenting MRI head anatomy~\cite{Hirsch2021}. Producing anatomically plausible results is critical in ensuring the minimisation of false positives, and promotes the clinical adoption of the developed models.

\subsubsection{Lesion-aware data augmentation}
Data augmentation plays a crucial role in medical imaging by enhancing the robustness and generalization of machine learning models trained on limited datasets. In medical imaging, acquiring large, diverse datasets can be challenging due to scarcity of labelled data, privacy concerns, and the complexity of data acquisition processes. Traditional data augmentation methods, such as rotation, flipping, scaling, and adding noise, have become a de facto part of medical image segmentation. Recently, disease-specific data augmentation methods have been introduced to further improve abnormal tissue segmentation. Such augmentation techniques can simulate rare or pathological conditions, enabling models to better handle real-world scenarios where such cases may be encountered infrequently. CarveMix, a mix-based augmentation method for brain lesion images~\cite{Zhang2021}, enriches the diversity of training data by stochastically selecting lesion regions from one image and superimposing them onto a designated target image. LesionMix augments brain lesion images by performing either lesion populating or lesion inpainting~\cite{basaran2023}, which provides controllable lesion-level data augmentation with user-specified lesion loads. Generative data augmentation methods offer an alternative approach through the synthesis of abnormalities. Reinhold et al. generates lesions with a specific lesion load through the use of a structural causal model~\cite{Reinhold2021}. Basaran et al. applies a cyclic attention-based generative adversarial network for synthesising lesions on healthy images~\cite{Basaran2022pseudo}. 

\subsection{Contributions} 
A preliminary version of this work has been published in MICCAI 2024 Longitudinal Disease Tracking and Modelling Workshop~\cite{basaran2024seghed}, in which we proposed SegHeD, a method for segmenting lesions using heterogeneous data input. Here, we substantially extend the method, experiments and results discussion. We propose SegHeD+, a multi-task segmentation model that can segment all, new, and vanishing lesions by learning from heterogeneous datasets. Compared to SegHeD, SegHeD+ employs lesion-aware data augmentation with the aim to address issues of data sparsity and lesion type imbalance in model training. Experimental results show that SegHeD+ outperforms SegHeD in terms of Dice score and $F_1$-score for MS lesion segmentation. SegHeD+ also differs from another heterogeneous segmentation method~\cite{Wu2023} by addressing vanishing lesions, implementing multiple combinations of heterogeneous images and labels for input, and proposing volumetric and spatial constraints. Overall, SegHeD+ demonstrates promising results in all and new lesion segmentation, and sets a benchmark for the overlooked task of vanishing lesion segmentation. The main contributions of this work are summarised as follows:

\begin{itemize}
\item We offer a comprehensive framework that takes into consideration diverse data types and annotation protocols, accounting for cross-sectional and longitudinal images, and all, new, and vanishing lesion labels.

\item We tackle various tasks concurrently, while most state-of-the-art brain lesion segmentation methods take a homogeneous segmentation approach, segmenting a single type of lesions. In addition to all and new lesions, we also account for vanishing (disappearing) lesions, an aspect overlooked in previous studies.

\item We incorporate domain knowledge about MS lesion segmentation via a novel combination of losses, accounting for the longitudinal relation between newly forming and vanishing lesions between timepoints, the volumetric change of lesions, and constraining lesion segmentation to anatomically plausible regions.

\item We utilise a lesion-level augmentation method to expand the training dataset, which is able to augment both cross-sectional and longitudinal images with new label availabilities.
\end{itemize}

\section{METHOD}
\subsection{Overall architecture}
\begin{figure*}[h!] \centering
\includegraphics[width=\textwidth]{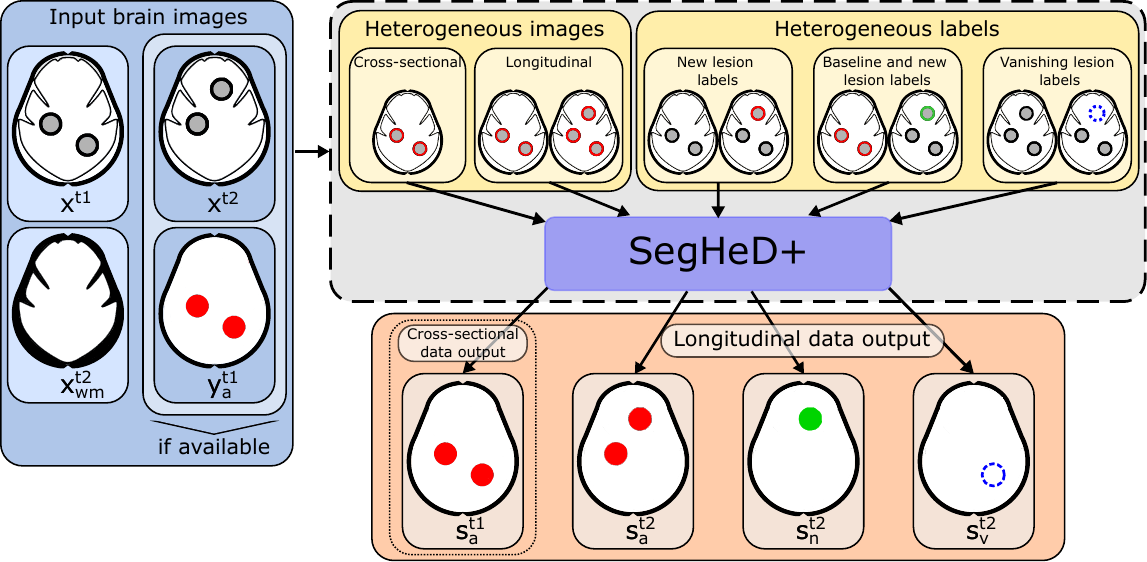}
\caption{Visualisation of the proposed framework. SegHeD+ learns from heterogeneous datasets varying in image and label formats. SegHeD+ takes up to four inputs, and can analyse cross-sectional and longitudinal data. Four segmentation heads provide binary-class segmentation to all lesions (red) in two timepoints, and new (green) and vanishing (dashed blue) lesions in a second timepoint.} 
\label{fig:framework}
\end{figure*}

SegHeD+ is a versatile model designed to learn from heterogeneous MS imaging data. Regarding data heterogeneity, it considers two types of data formats: cross-sectional and longitudinal. Regarding label heterogeneity, the model accommodates three annotation protocols: all-lesion annotation, new-lesion annotation (which annotates only new lesions in the second scan), and vanishing-lesion annotation (which annotates lesions that disappear in the second scan). The latter two protocols concentrate on the longitudinal changes of lesions and refrain from annotating all lesions to optimize time efficiency. SegHeD+ is designed to tackle all three annotation tasks simultaneously. The overall framework of the method is depicted in Figure~\ref{fig:framework}.

We make the assumption that the imaging data consists of two timepoints. SegHeD+ requires 4 input channels: a scan at the first timepoint (baseline), $x^{t1}$, a scan at the second timepoint (follow-up), $x^{t2}$, a label map of all lesions for the first timepoint scan, $y^{t1}_{a}$, and a white matter mask for the second timepoint scan, $x^{t2}_{wm}$. In cases of heterogeneous datasets, not all of the aforementioned inputs may be present. If data for the second timepoint is missing, the first timepoint data is duplicated as the second, $x^{t2} = x^{t1}$. If the label map for all lesions at the first timepoint is unavailable, $y^{t1}_{a}$ is substituted with a zero matrix of the same dimensions as $x^{t1}$. The white matter mask for the second timepoint, $x_{wm}^{t2}$, can be acquired using a pre-trained brain structure segmentation tool, such as SynthSeg, which is robust in white matter segmentation even in the presence of lesions~\cite{Billot2023}. SegHeD+ is trained to produce 4 outputs: segmentation of all lesions at the first timepoint, $s_{a}^{t1}$, segmentation of all lesions at the second timepoint, $s_{a}^{t2}$, segmentation of new lesions at the second timepoint, $s_{n}^{t2}$, and segmentation of vanishing lesions at the second timepoint, $s_{v}^{t2}$. The segmentation process is formulated by
\begin{equation}
\label{eq:form}
    s_{a}^{t1}, s_{a}^{t2}, s_{n}^{t2}, s_{v}^{t2} =  F(x^{t1}, x^{t2}, y_{a}^{t1}, x_{wm}^{t2}),
\end{equation}
where $F$ represents the SegHeD+ model. The all-lesion map, $y^{t1}_{a}$, is not utilised in cases of cross-sectional data input. It is only input, when available, to provide temporal context for second timepoint lesion prediction, and not used for predicting $s_{a}^{t1}$. 

Due to computational constraints, the model handles a maximum of two timepoints as inputs. In cases where a subject has longitudinal scans with more than two timepoints~\cite{Carass2015}, a sliding window technique is applied across the timepoints during training and inference. For example, for case with three timepoints, [t1, t2, t3], we successively provide two timepoints as input to the model, formulated as [t1, t1], [t1, t2], [t2, t3]. The model first processes images at [t1, t1], i.e. using two channels of identical images as input. After the segmentation output for t1 is computed,  the model processes images at [t1, t2], so that the previous timepoint t1 can provide temporal information in processing t2. After the output for t2 is computed, the model further processes two channels of images at [t2, t3]. During inference for longitudinal images, the all-lesion segmentation of a timepoint, for example $s_{a}^{t1}$, is used as an input channel, representing $y^{t1}_{a}$, for temporal guidance to obtain the second timepoint all-lesion segmentation $s_{a}^{t2}$. SegHeD+ utilises a 3D V-Net architecture~\cite{Milletari2016} and a combined training loss function that incorporates the Dice loss, a longitudinal constraint loss, a spatial constraint loss, and a volumetric constraint loss.

\subsection{Anatomical constraints}
SegHeD+ is trained using a mix of loss functions, which we term as \textit{anatomical constraints}. This approach is inspired by the way radiologists examine longitudinal images to detect lesions~\cite{Homssi2023}. Anatomical constraints formulate the relationships in terms of time, space, and volume for segmenting brain lesions at two different time points.

\subsubsection{Longitudinal constraints}
We utilise the existing information embedded in the lesion annotation protocol: (1) Predicted new-lesions at the second timepoint, denoted as $s_{n}^{t2}$, should not exist in the all-lesion label at the initial timepoint, $y_{a}^{t1}$, but must be included in the all-lesion label at the second timepoint, $y_{a}^{t2}$. (2) Predicted disappearing lesions at the second timepoint, $s_{v}^{t2}$, must be present in the all-lesion label at the first timepoint, $y_{a}^{t1}$, but should not be present in the all-lesion label at the second timepoint, $y_{a}^{t2}$. These longitudinal constraints are formulated using a mean squared error loss,
\begin{equation}
\label{eq:long}
\begin{aligned}
\mathcal{L}_{Long}=\frac{1}{\text{N}} \sum_{i=1}^{\text{N}} \|y^{t1}_{a_{i}} \land p^{t2}_{n_{i}}\|^2 + 
\frac{1}{\text{N}} \sum_{i=1}^{\text{N}} \|y^{t2}_{a_{i}} \oplus p^{t2}_{n_{i}}\|^2 + \\ 
\frac{1}{\text{N}} \sum_{i=1}^{\text{N}} \|y^{t1}_{a_{i}}\oplus p^{t2}_{v_{i}}\|^2 + 
\frac{1}{\text{N}} \sum_{i=1}^{\text{N}} \|y^{t2}_{a_{i}}\land p^{t2}_{v_{i}}\|^2,
\end{aligned}
\end{equation}
where $\text{N}$ represents the total number of training images, $i$ denotes the index of the image, $\land$ represents the element-wise logical AND operation, $\oplus$ denotes the element-wise logical XOR operation, and $|| \cdot|| ^2$ represents the squared L2-norm of a flattened image. The superscripts, [t1, t2], for the prediction, $p$, and label, $y$, denote the two timepoints used by the sliding window. In contrast to the regularisation proposed in~\cite{Wu2023}, we apply longitudinal constraints to predictions of new and vanishing lesions rather than all-lesion predictions. In cases where datasets are cross-sectional, longitudinal constraints are not enforced.

\subsubsection{Volumetric constraints}
\label{sec:volconst}
Changes in brain lesion volumes evolve over time. Lesions can grow, shrink, appear, and/or vanish. Both change in lesion volume and lesion count provide important biomarkers for disease progression~\cite{Filippi1995, Filippi2010, Tiu2022}. Unfortunately, a subject with MS can have both hundreds of small lesions or few large lesions, showing high variation in lesion count per subject and over time. This variation can be addressed by measuring total lesion volume change, also reducing the need for three-dimensional connected component analysis during training. To address the correlation between lesion volumes at two different timepoints, a volumetric constraint loss is proposed. This loss penalises significant discrepancies in lesion volume. It permits variations in lesion volume to either rise or fall~\cite{Sethi2017,Genovese2019, Filippi1995} within specified thresholds, denoted as $\alpha_{high}$ and $\alpha_{low}$. The penalty is enforced only when the volume alteration exceeds these predefined thresholds. The volumetric constraint loss is formulated as
\begin{equation}
\label{eq:volumetric}
\mathcal{L}_{Vol} = \begin{cases}
  \frac{1}{\text{N}} \sum_{i=1}^{\text{N}}(V_{a_{i}}^{t2} - \alpha_{high} \cdot V_{a_{i}}^{t1})^2 & \text{if } V_{a_{i}}^{t2} \geq \alpha_{high} \cdot V_{a_{i}}^{t1} \\
  \frac{1}{\text{N}} \sum_{i=1}^{\text{N}}(V_{a_{i}}^{t2} - \alpha_{low} \cdot V_{a_{i}}^{t1})^2   & \text{if } V_{a_{i}}^{t2}  \leq \alpha_{low} \cdot V_{a_{i}}^{t1} \\
  0   & \text{otherwise}, 
\end{cases}
\end{equation}
where $V_{a_{i}}^{t1}$ and $V_{a_{i}}^{t2}$ represent the total volume of lesions at the first timepoint and second timepoint for the $i$-th image, respectively. Lesion volumes are calculated by taking a sum of the lesion maps, as 0's represent healthy, and 1's represent lesion tissues. To our knowledge, there is no research on the extent to which lesion volume might \textit{decrease} over time. We determine $\alpha_{low}$ to be 0.8 based on an examination of the largest reduction in lesion volume in a longitudinal MS dataset~\cite{Carass2015}. Likewise, using the same MS dataset, we established $\alpha_{high}$ to be 1.2. These values are in line with existing medical literature~\cite{Molyneux1998} for annual rate of white matter lesion change. Therefore, values $\alpha_{high}$ and $\alpha_{low}$ are defined for the annual rate of change, and can be modified when using for subjects that are scanned at a non-annual rate. A sensitivity study on $\alpha_{high}$ and $\alpha_{low}$ supported setting these hyperparameters to 1.2 and 0.8, respectively. When working with a cross-sectional dataset, as input $x^{t2} = x^{t1}$, we expect outputs $s_{a}^{t1}$ and $s_{a}^{t2}$ to be equal. As such, in cases of cross-sectional data input, we set $\alpha_{high}$ and $\alpha_{low}$ to 1 in order to guide the network to generate identical outputs. 

\subsubsection{Spatial constraint}
MS affects the central nervous system of the body, where lesions may present themselves in the spinal cord, white matter, and rarely in the gray matter of the brain~\cite{Filippi2010}, alongside leading to gray matter atrophy. Studies demonstrate cognitive impairment in neurodegenerative diseases is significantly associated to white matter lesions in the brain~\cite{Tiu2022}. As such, we do not consider gray matter atrophy, or lesions in the spinal cord during training or inference. Additionally, the spinal cord is excluded in most MS datasets. White matter lesions appear as hyperintense regions on FLAIR MR images~\cite{Lassmann2018}. To integrate this information into our framework, we adopt two approaches: first, by utilizing the white matter mask, $x_{wm}^{t2}$, as an input channel for SegHeD+ to benefit from the information of anatomical structure; and secondly, by creating a spatial relation loss function, where we further penalises false positive segmentations which lie outside of the white matter region. This function is defined as the mean squared error between the segmented lesions, $s$, which are located outside the white matter, $x_{wm}^{t2}$, and a zero matrix of the same dimensions as the image:
\begin{equation}
\label{eq:spatial}
\mathcal{L}_{Spat}=\frac{1}{\text{N}} \sum_{i=1}^{\text{N}} \|p \oplus (\mathbf{1}-x_{wm_{i}}^{t2})\|^2.
\end{equation}

The total loss, $\mathcal{L}$, is defined as
\begin{equation}
\begin{aligned}
\label{eq:total}
\mathcal{L} = \begin{cases}
  \mathcal{L}_{Dice} & \text{if } n < \dfrac{n_{epochs}}{2},\\
  \mathcal{L}_{Dice} + \lambda_{L} \cdot \mathcal{L}_{Long} + \\
  \indent \lambda_{V} \cdot \mathcal{L}_{Vol} + \lambda_{S} \cdot\mathcal{L}_{Spat}  & \text{if } n \geq \dfrac{n_{epochs}}{2},
\end{cases}
\end{aligned}
\end{equation}
where $\mathcal{L}_{Dice}$ represents the Dice loss, $n$ stands for the epoch, $n_{epochs}$ represents the total number of epochs for training, and $\lambda_{Long}, \lambda_{Vol}$, and $\lambda_{Spat}$ are parameters used for weighting. We employ a curriculum learning strategy~\cite{Bengio2009} and introduce the complete loss after half the number of epochs.

\subsection{Lesion-aware data augmentation}
Apart from learning from heterogenous datasets, we also enlarge the training set and lesion diversity by performing data augmentation. On top of traditional data augmentation techniques, we introduce LesionMix, a lesion-aware data augmentation method~\cite{basaran2023}. LesionMix operates at the lesion level, enabling increased diversity in lesion shape, location, intensity, and load distribution. It  can perform both lesion populating, which increases lesion load by mixing an input image with augmented lesions, and lesion inpainting, which decreases lesion load by inpainting randomly selected lesions. Lesion populated images are produced by
\begin{equation}
\label{eq:mix1}
    \text{X'}   =  \text{X} \odot (1-\text{M}) + \text{F} \odot \text{M}
\end{equation}
\begin{equation}
\label{eq:mix2}
    \text{Y'} =  \text{Y} \odot (1 -\text{M}) +  \text{M}, 
\end{equation}
where $\text{X}$ is the input image, $\text{Y}$ is its lesion mask, $\text{F}$ denotes the augmented lesion intensity image, $\text{M}$ denotes the mask of the augmented lesion region, and $\odot$ denotes element-wise multiplication. $\text{F}$ and $\text{M}$ are drawn from a pool of real lesion samples, followed by lesion-level augmentation including shape augmentation, intensity manipulation, and Gaussian noise addition to increase diversity as described in~\cite{basaran2023}.

Lesion inpainted images are produced by
\begin{equation}
\label{eq:inpaint1}
    \text{X'} = G(f(\text{X}, \text{M}))\odot \partial \text{M} + f(\text{X}, \text{M}) \odot (1-\partial \text{M})
\end{equation}
\begin{equation}
\label{eq:inpaint2}
    \text{Y'} =  \text{Y} -  \text{M},
\end{equation}
where $f(\text{X}, \text{M})$ denotes the inpainting function using the fast marching algorithm~\cite{Telea2004}, $G$ denotes the Gaussian blurring function to ensure a smooth boundary, and $\partial \text{M}$ denotes the boundary of the lesion mask. \ref{eq:inpaint1} inpaints $\partial \text{M}$ and is performed iteratively until a predetermined lesion volume distribution is achieved~\cite{basaran2023}.

\begin{figure}[!t] \centering
\includegraphics[width=.8\linewidth]{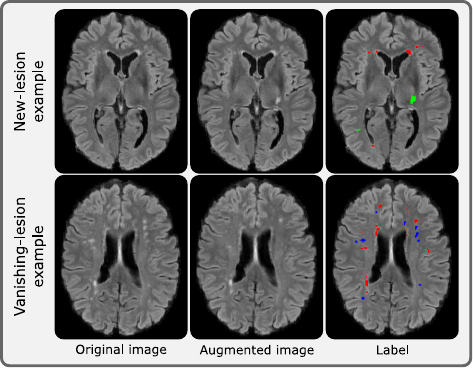}
\caption{Example outputs of LesionMix augmentation. Red labels denote lesions which are present in the original image; green labels denote new-lesion generated in the augmented image; blue labels denote vanishing-lesions which have been inpainted from the original image. Images best viewed online.} 
\label{fig:lesionmix}
\end{figure}

LesionMix was originally proposed for augmenting cross-sectional brain images. Here we employ it to augment both cross-sectional and longitudinal brain images. LesionMix complements SegHeD+ in three manners. First, it can generate both new lesions and vanishing lesions, thus increasing the diversity of the the various labels SegHeD+ can analyse in heterogeneous datasets. This both increases dataset variance, and reduces the effect of label imbalances which may occur from one type (new lesions) appearing more than another one (vanishing lesions). Second, its lesion populating and inpainting strategy along with lesion mask output allows for synthetic generation of longitudinal data. Lesions can be populated at the second timepoint as new lesions, or inpainted as vanishing lesions. Finally, LesionMix operates by populating or inpainting lesions to a predetermined lesion load, thus can control lesion volume for augmented images~\cite{basaran2023}. By integrating LesionMix with the volumetric constraint knowledge and uniformly sampling between the lesion volume increase and decrease thresholds, $\alpha_{high}$ and $\alpha_{low}$, we set a target load for an image, and augment it to generate an augmented second timepoint. We refer the reader to ~\cite{basaran2023} for more implementation details of LesionMix. Here, we present example outputs of LesionMix augmentation in Figure~\ref{fig:lesionmix}. 

\section{EXPERIMENTS}
\subsection{Datasets}
We curate five datasets of brain lesions from MS patients that exhibit heterogeneities in data format and annotation. The first three datasets are publicly available. For all datasets, we utilise the FLAIR modality as it is the common modality among the datasets.

\indent\textbf{ISBI 2015 MS lesion dataset} (MS2015)~\cite{Carass2015} is a public longitudinal MS brain MRI dataset. The training set consists of 5 subjects, four of which with four timepoints, and the fifth subject with five timepoints, totaling 21 scans with respective annotations for all-lesions. The annotations of the test set are not publicly available. We split the original training set of 21 images into a training subset of 13 images, and a test subset of 8 images. All subjects are scanned using the same 3T Philips Medical Systems scanner, with follow-up scans acquired at a mean of 1 year intervals. The dataset is provided preprocessed with inhomogeneity correction using N4, skull-stripping, dura-stripping, followed by a second N4 inhomogeneity correction, and rigid registration to 1mm isotropic MNI template. 

\indent\textbf{MICCAI 2016 MS lesion dataset} (MS2016)~\cite{Commowick2018} is a public cross-sectional MS brain MRI dataset. The training dataset comprises of 15 subjects, with respective annotations for all-lesions. The test set is not publicly available with annotations. We split the original training set of 15 images into a training subset of 11 images, and a test subset of 4 images. Subjects are scanned using a variety of scanners, including five subjects from Siemens Verio 3T, five subjects from Siemens Aera 1.5T, and five subjects from Philips Ingenia 3T MRI machines. A pre-processed dataset is provided by the organisers, which includes denoising using non-local means algorithm~\cite{Coupe2008}, rigid-registration onto the FLAIR image via block-matching~\cite{Commowick2012}, brain extraction using the volBrain platform~\cite{Manjon2016}, and bias field correction using the N4 algorithm~\cite{Tustison2010}.

\indent\textbf{MICCAI 2021 MS new-lesion segmentation challenge dataset} (MSSEG-2)~\cite{Commowick2021} is a public longitudinal MS brain MRI dataset. The training and testing datasets comprises of 40 and 60 subjects, respectively. Each subject is scanned at two timepoints, with only new-lesion annotations in the second timepoint to save the annotation cost. Second timepoint scans are images 1 to 3 years after the first one. A total of 15 different MRI scanners are represented, including 1.5T and 3T machines from General Electric, Philips, and Siemens MRI machines. The Anima-based\footnote{Anima scripts: RRID:SCR\_017072, https://anima.irisa.fr} MSSEG-2 longitudinal pre-processing script is implemented, which includes an atlas-based brain extraction tool~\cite{Doshi2013}, followed by N4 bias field correction~\cite{Tustison2010}.

\indent\textbf{MSSEG-2+} is a private longitudinal MS brain MRI dataset which is derived from MSSEG-2. It follows the same images, dataset structure and pre-processing protocol as MSSEG-2. MSSEG-2+ comprises of all-lesion annotations in the first timepoint, provided by an expert using ITK-SNAP~\cite{ITKSNAP}, along with the annotations for new lesions at the second timepoint. 

\indent\textbf{VAN} is a private longitudinal MS brain MRI dataset derived from MSSEG-2. In order to address the lack of public datasets for vanishing lesions, VAN is created by simulating vanishing lesions by the inversion of timepoints in the MSSEG-2 dataset. Consequently, new lesions at the second timepoint are transformed into lesions that disappear from the first timepoint. The timepoint inversion is done after the pre-processing pipeline defined in the MSSEG-2 dataset description.

These datasets consist of lesion images obtained from diverse MR scanners with varying annotation procedures. Currently, a longitudinal dataset with annotated all, new, and vanishing lesions does not exist. Therefore, a split from each dataset is held-out for testing and not used during training. In total, we benchmark our method using 192 FLAIR MR images. The high variance in medical scanners, voxel spacing, and image sizes within both the training and testing datasets diminishes cross-domain effects. Additional information on the datasets, including the division of training and test sets used, is available in Table~\ref{table:dataset}. During preprocessing, all images are resampled to a voxel spacing of $1\times1\times1$ mm$^3$, followed by white matter extraction using SynthSeg~\cite{Billot2023} and rigid registration to the MNI template space~\cite{Fonov2009}.

\begin{table}[htbp]
\begin{center}
\caption{Summary of the five datasets used for model training and evaluation, which contains different data formats (Image availability) and annotation styles (Label availability). $Y_{a}^{t1}$: first timepoint all lesion labels, $Y_{a}^{t2}$: second timepoint all-lesion labels, $Y_{n}^{t2}$: second timepoint new-lesion labels, $Y_{v}^{t2}$: second timepoint vanishing-lesion labels.}
\setlength{\tabcolsep}{1.4mm}{
\centering
\begin{tabular}[b]{lcccccccc}
\toprule
\multicolumn{3}{c}{Dataset} &\multicolumn{2}{c}{Image availability} & \multicolumn{4}{c}{Label availability} \\
\cmidrule(r){1-3}\cmidrule(lr){4-5}\cmidrule(l){6-9}
Name & Train images & Test images & Cross-sectional & Longitudinal & $Y_{a}^{t1}$ & $Y_{a}^{t2}$ & $Y_{n}^{t2}$ & $Y_{v}^{t2}$ \\
\midrule
MS2015 & 13 & 8 & - &\checkmark &\checkmark &\checkmark & - & -  \\
MS2016 & 11 & 4 & \checkmark & - & \checkmark & - & - & -  \\
MSSEG-2 & 40 & 60 & - &\checkmark & - & - & \checkmark & - \\
MSSEG-2+  & 40 & 60 & -  & \checkmark& \checkmark & - & \checkmark & - \\
VAN  & 40 & 60 & - &\checkmark& - & - & - & \checkmark \\
\midrule
Total & 144 & 192 &&&&&&\\
\cmidrule[.75pt](r){1-3}
\label{table:dataset}
\end{tabular}}
\end{center}
\end{table}
\vspace{-1cm}

\subsection{Implementation details}
\subsubsection{Network and hyperparameters}
SegHeD+ is constructed using a 3D V-Net architecture~\cite{Milletari2016} with five downsampling layers, and four heads in the final layer for the four lesion segmentation tasks. It is implemented using PyTorch 2.2 and trained on Nvidia Tesla T4 GPUs. Training and inference is performed with an image patch size of $96\times96\times96$, which shows optimal performance taking into consideration GPU constraints. The model is trained from scratch using the Adam optimizer with a learning rate of 0.001 for 20,000 epochs. The hyperparameters $\lambda_{L}$, $\lambda_{V}$, and $\lambda_{S}$ were empirically set to 2, 1, and 1, respectively.

An important consideration in the network is the aid of the all-lesion label, $y_{a}^{t1}$ in guiding temporal segmentation. As mentioned, this label is not utilised for cross-sectional input. For temporal segmentation tasks, such as in MS2015, we provide this input 50\% of the the time during training for longitudinal data. This ensures that the network does not become overly dependent on the label for guiding the second timepoint all-lesion segmentation, $s_{a}^{t2}$. For testing, we report results for when $y_{a}^{t1}$ is, and is not input for longitudinal inference. 

\subsubsection{Data augmentation}
\begin{figure}[!h] \centering
\includegraphics[width=.9\linewidth]{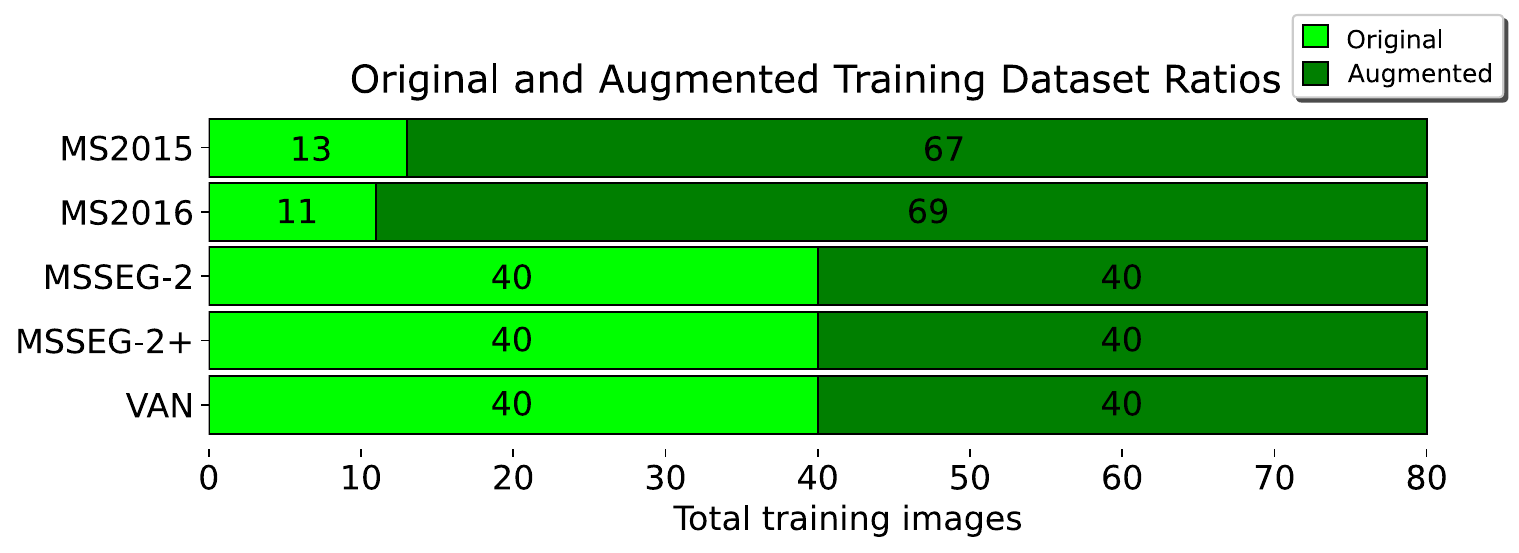}
\caption{Visual illustration of the size of original training data and LesionMix-augmented training data.} 
\label{fig:stackedbar}
\end{figure}

Data augmentation is a crucial element in the SegHeD+ pipeline. We perform offline lesion-level augmentation, LesionMix, followed by on-the-fly traditional data augmentations during model training. We implement LesionMix such that we equalise the training data imbalance in the different datasets in Table~\ref{table:dataset}, and increase the size of overall training data. This is essential, as the dataset imbalance also impacts the overall testing performance for different segmentation tasks. For example, three of the datasets provide diverse images for all-lesion segmentation, however, only one dataset provides images for vanishing-lesion segmentation. While the model becomes more robust and generalisable for the all-lesion segmentation task, it may suffer in the latter task. Prior to training, we perform offline LesionMix augmentation using its default parameters as described in~\cite{basaran2023} with lesion volume increase and decrease thresholds $\alpha_{high}$ and $\alpha_{low}$ for longitudinal images, defined in section~\ref{sec:volconst}. Table~\ref{table:lesionmix_dataset} reports the training set size after LesionMix is applied. Figure~\ref{fig:stackedbar} portrays the balance between original datasets and LesionMix generated datasets.

\begin{table}[htbp]
\begin{center}
\caption{Summary of the augmented five datasets, after LesionMix is applied. Note that during model training, traditional data augmentation will be further applied on-the-fly. $Y_{a}^{t1}$: first timepoint all lesion labels, $Y_{a}^{t2}$: second timepoint all-lesion labels, $Y_{n}^{t2}$: second timepoint new-lesion labels, $Y_{v}^{t2}$: second timepoint vanishing-lesion labels.}
\setlength{\tabcolsep}{1.4mm}{
\centering
\begin{tabular}[b]{lcccccccc}
\toprule
\multicolumn{3}{c}{Dataset} &\multicolumn{2}{c}{Image availability} & \multicolumn{4}{c}{Label availability} \\
\cmidrule(r){1-3}\cmidrule(lr){4-5}\cmidrule(l){6-9}
Name & Train images & Test images & Cross-sectional & Longitudinal & $Y_{a}^{t1}$ & $Y_{a}^{t2}$ & $Y_{n}^{t2}$ & $Y_{v}^{t2}$ \\
\midrule
MS2015 & 80 & 8 & - &\checkmark &\checkmark &\checkmark & \checkmark & \checkmark  \\
MS2016 & 80 & 4 & \checkmark & \checkmark & \checkmark & \checkmark & \checkmark & \checkmark  \\
MSSEG-2 & 80 & 60 & - &\checkmark & - & - & \checkmark & \checkmark \\
MSSEG-2+  & 80 & 60 & -  & \checkmark& \checkmark & \checkmark & \checkmark & \checkmark \\
VAN  & 80 & 60 & - &\checkmark& - & - & \checkmark & \checkmark \\
\midrule
Total & 400 & 192 &&&&&&\\
\cmidrule[.75pt](r){1-3}
\label{table:lesionmix_dataset}
\end{tabular}}
\end{center}
\end{table}
\vspace{-2cm}

During training, we implement on-the-fly traditional data augmentation methods, including flipping and rotation in three dimensions, elastic deformation, brightness adjustments (additive and multiplicative), and additive Gaussian noise using the \textit{batchgenerators} framework\footnote{\url{https://github.com/MIC-DKFZ/batchgenerators}}. Five-fold cross-validation is performed, creating an ensemble of five trained models. After training, for each fold, we select the model which produces the highest Dice score. Lesion segmentation performance is evaluated on the test set using an ensemble of the five trained models by averaging their softmax outputs before binary classification for each segmentation head.

\subsection{Evaluation metrics}
We evaluate the performance of SegHeD+ in terms of lesion segmentation, measuring segmentation overlap, and lesion-wise detection. Lesion segmentation performance is evaluated using the Dice similarity coefficient, formulated as
\begin{equation}
    \text{Dice} = 2\frac{\mid A\cap G\mid}{\mid A\mid + \mid G\mid},
\end{equation}
where A represents the automated segmentation and G represents the ground truth segmentation. The ideal Dice score is 1. The performance of lesion detection is assessed using the lesion-wise $F_1$ score, which quantifies the accuracy of lesion detection irrespective of the precision of their boundaries. The $F_1$ score is calculated as
\begin{equation}
    F_1 = 2 \frac{S_L \cdot P_L}{S_L + P_L},
\end{equation}
where $S_{L}$ represents the sensitivity of lesion detection (true positive rate or recall) and $P_{L}$ represents the precision of positive predictive value. The ideal $F_{1}$ score is 1. A lesion is classified as detected (true positive) when the automated detection intersects with a minimum of 10\% of the ground truth lesion volume and does not exceed 70\% of the volume, following the practice in~\cite{Commowick2018}. Using these metrics and parameters allows the direct comparison of SegHeD+ to the MSSEG-2 challenge competitors.

The Dice and $F_1$ scores are calculated using the \texttt{animaSegPerfAnalyzer2} function of the Anima analyzer tool\footnote{Anima scripts: RRID:SCR\_017072, https://anima.irisa.fr}, as provided by the organizers of the MSSEG-2 challenge to ensure a fair comparison with other methods. Post-processing is performed using the default settings of animaSegPerfAnalyzer, which filters out lesion volumes smaller than $3\text{mm}^3$. This is consistent with the MSSEG-2 challenge evaluation protocol~\cite{Basaran2022, Kamraoui2022}, for fair comparison to challenge competitors.

\subsection{Results}
\begin{figure*}[!t] \centering
\includegraphics[width=.9\textwidth]{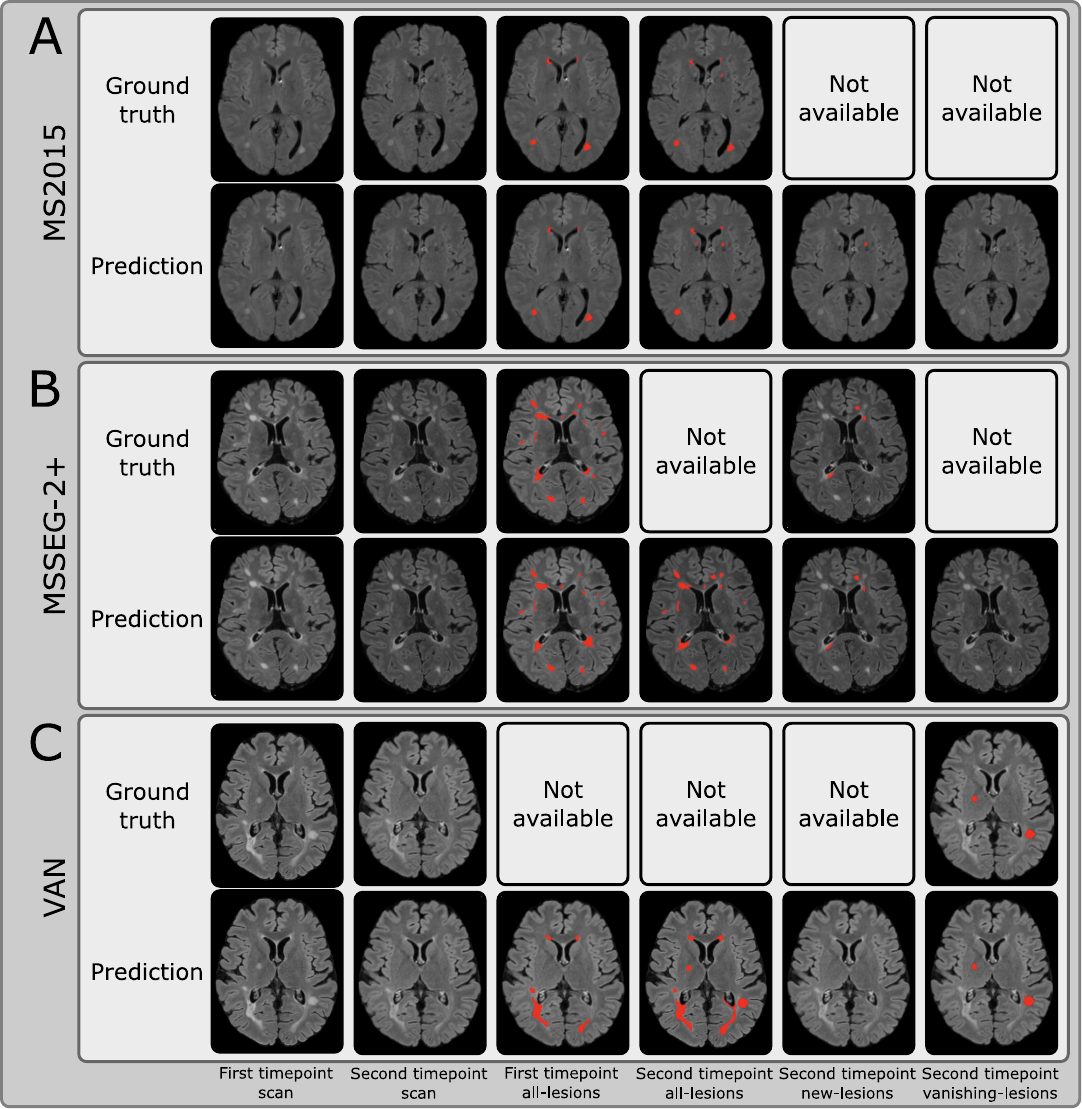}
\caption{SegHeD+ is capable of simultaneous multi-task segmentation (Rows 3 to 6). Some tasks do not show new/vanishing-lesions predictions as they are not present at the given slice. ``Not available" denotes no ground truth annotation for comparison. \textbf{A:} Dataset where all-lesion labels are available for first and second timepoints. \textbf{B:} Dataset where first timepoint all-lesion label and second timepoint new-lesion label are available. \textbf{C:} Dataset where second timepoint vanishing-lesion label is available.} 
\label{fig:segmentations2}
\end{figure*}

SegHeD+ is a model capable of performing three segmentation tasks, namely all-lesion, new lesion and vanishing lesion segmentations. We assess its performance across multiple tasks and compare it with state-of-the-art task-specific segmentation approaches such as nnU-net~\cite{Isensee2021}, nnFormer~\cite{Zhou2023}, UNETR~\cite{Hatamizadeh2022}, a recent method for learning from heterogeneous data called CoactSeg~\cite{Wu2023}, a previous iteration of this work, SegHeD~\cite{basaran2024seghed}, which does not feature LesionMix augmentation, and methods specifically tailored for new lesion segmentation~\cite{Zhang2021, Basaran2022}, as presented in Table~\ref{table:segresults}. It has to be noted that only SegHeD and SegHeD+ are capable of performing all three tasks simultaneously. The task-specific SOTA methods undergo two rounds of training: one for the all-lesion segmentation task using the MS2015 and MS2016 datasets, and another for the new-lesion segmentation task using the MSSEG-2 dataset. CoactSeg~\cite{Wu2023} is trained only once, as it can handle data input and inference for all-lesion and new-lesion segmentation. For new lesion segmentation with the MSSEG-2 dataset, we present Dice scores for MedICL~\cite{zhang2021msseg}, Basaran~\cite{Basaran2022}, and the average score from four human experts, denoted as ``Avg. of Experts", officially reported by the challenge organizers~\cite{Commowick2021}. The ``Avg. of Experts" can be regarded as the upper bound performance for new-lesion segmentation on MSSEG-2. Additionally, we include lesion-wise $F_1$ scores in Table~\ref{table:f1}, as per the guidelines of the MSSEG-2 challenge. Example segmentations are compared in Figure~\ref{fig:segmentations1}, with more examples provided in the Figure~\ref{fig:segmentations2}.

\begin{table}[h] \centering
\setlength{\tabcolsep}{.5em}
\begin{center}
\caption{Mean and standard deviations of lesion segmentation Dice scores ($\%$). Best results are in bold. N/A: output not available for the given method. $\star$ indicates that $y_{a}^{t1}$ was used for longitudinal all-lesion segmentation. $\dagger$: Methods where two models need to be trained, one for all-lesion and one for new-lesion segmentation. $\ddagger$: Not trained on MSSEG-2+. Asterisks indicate statistical significance  ($^{*}$:~p$\leq$~0.05, $^{**}$:~p~$\leq$ 0.01, $^{***}$:~p~$\leq$ 0.005) when using a paired Student's \textit{t}-test comparing SegHeD+'s performance to benchmarked methods.}
\label{table:segresults}
\begin{tabular}[b]{p{23mm} lccccc} \toprule
             & & \multicolumn{5}{c}{Segmentation task (lesion type)}\\
             \cmidrule(lr){3-7} 
             &  & \multicolumn{2}{c}{All} & \multicolumn{1}{c}{New} &\multicolumn{1}{c}{All\&New}  & \multicolumn{1}{c}{Vanishing} \\ 
            \cmidrule(lr){3-4} \cmidrule(lr){5-5} \cmidrule(lr){6-6} \cmidrule(lr){7-7} 
            Type & Method & \multicolumn{1}{c}{MS2015} & \multicolumn{1}{c}{MS2016} &\multicolumn{1}{c}{MSSEG-2}  & \multicolumn{1}{c}{MSSEG-2+}  & \multicolumn{1}{c}{VAN} \\ 
     \cmidrule(lr){1-2} \cmidrule(lr){3-3} \cmidrule(lr){4-4} \cmidrule(lr){5-5} \cmidrule(lr){6-6} \cmidrule(lr){7-7} 
        \multirow{4}{*}{\shortstack[l]{New-\\ lesion\\methods}} & MedICL~\cite{zhang2021msseg} & N/A & N/A & $50.67_{29.38}^{}$ & N/A & N/A \\ 
        &Basaran et al.~\cite{Basaran2022}  & N/A & N/A & $51.06_{28.92}^{}$ & N/A & N/A \\ 
        &LaBRI-IQDA~\cite{Commowick2021}  & N/A & N/A & $49.82_{29.96}$ & N/A & N/A \\ 
        &Avg. of Experts~\cite{Commowick2021}  & N/A & N/A & $\mathbf{55.52_{34.43}}$ & N/A  & N/A\\\midrule
        \multirow{5}{*}{\shortstack[l]{Task-\\ specific\\ SOTA}}&nnU-Net~\cite{Isensee2021} $\dagger$ & $73.01_{4.91}^{***}$ & $74.87_{7.54}^{***}$ & \multicolumn{1}{|c}{$48.89_{31.20}^{}$} & N/A & N/A \\ 
        &nnFormer~\cite{Zhou2023} $\dagger$ & $72.56_{7.15}^{***}$ & $74.12_{8.52}^{***}$ & \multicolumn{1}{|c}{$47.01_{33.39}^{*}$} & N/A & N/A \\
        &UNETR~\cite{Hatamizadeh2022} $\dagger$ & $72.79_{6.13}^{***}$ & $73.78_{7.98}^{***}$ & \multicolumn{1}{|c}{$45.51_{30.84}^{**}$} & N/A & N/A \\
        &TransBTS~\cite{transbts} $\dagger$ & $72.60_{8.00}^{***}$ & $74.03_{8.34}^{***}$ & \multicolumn{1}{|c}{$46.58_{30.07}^{*}$} & N/A & N/A \\
        & TransUNet~\cite{Chen2021TransUNetTM} $\dagger$ & $70.98_{7.29}^{***}$ & $73.15_{8.99}^{***}$ & \multicolumn{1}{|c}{$41.48_{33.32}^{***}$} & N/A & N/A \\ \midrule
        \multirow{3}{*}{\shortstack[l]{Hetero-\\geneous\\methods}}&CoactSeg~\cite{Wu2023} $\ddagger$  & $71.28_{8.24}^{***}$ & $71.31_{9.15}^{***}$ & $47.35_{38.12}^{}$ & $58.54_{18.54}^{***}$ & N/A \\
        &SegHeD~\cite{basaran2024seghed} & $74.87_{6.13}$ & $84.73_{7.12}$ &$48.64_{33.81}$ & $65.51_{19.67}$ & $35.23_{20.62}$ \\
        &\textbf{SegHeD+} & $\mathbf{78.57_{6.71}}$ & $\mathbf{85.18_{7.10}}$ &$50.52_{31.29}$ & $\mathbf{66.83_{18.98}}$ & $\mathbf{43.84_{18.04}}$ \\ \bottomrule
\end{tabular}
\end{center}
\end{table}

\begin{table}[h] \centering
\setlength{\tabcolsep}{.5em}
\caption{Mean and standard deviations of lesion detection $F_1$ scores ($\%$), in accordance with the MSSEG-2 challenge~\cite{Commowick2021}. Best results are in bold. $\star$ indicates that $y_{a}^{t1}$ was used for longitudinal all-lesion segmentation. $\dagger$: Methods where two models are trained, one for all lesion and one for new lesion segmentation. $\ddagger$: Not trained on MSSEG+. Asterisks indicate statistical significance  ($^{*}$:~p$\leq$~0.05, $^{**}$:~p~$\leq$ 0.01, $^{***}$:~p~$\leq$ 0.005) when using a paired Student's \textit{t}-test comparing SegHeD+ to competing methods.}
\label{table:f1}
\begin{tabular}[b]{p{23mm} lccccc} \toprule
            & & \multicolumn{5}{c}{Segmentation task (lesion type)}\\
             \cmidrule(lr){3-7} 
             &  & \multicolumn{2}{c}{All} & \multicolumn{1}{c}{New} &\multicolumn{1}{c}{All\&New}  & \multicolumn{1}{c}{Vanishing} \\ 
            \cmidrule(lr){3-4} \cmidrule(lr){5-5} \cmidrule(lr){6-6} \cmidrule(lr){7-7} 
            Type & Method & \multicolumn{1}{c}{MS2015} & \multicolumn{1}{c}{MS2016} &\multicolumn{1}{c}{MSSEG-2}  & \multicolumn{1}{c}{MSSEG-2+}  & \multicolumn{1}{c}{VAN} \\ 
            \cmidrule(lr){1-2} \cmidrule(lr){3-3} \cmidrule(lr){4-4} \cmidrule(lr){5-5} \cmidrule(lr){6-6} \cmidrule(lr){7-7} 
        \multirow{4}{*}{\shortstack[l]{New-\\ lesion\\methods}} & MedICL~\cite{zhang2021msseg} & N/A & N/A & $49.98_{32.78}^{*}$ & N/A & N/A \\ 
        & Basaran et al.~\cite{Basaran2022}  & N/A & N/A & $55.25_{34.81}^{}$ & N/A & N/A \\ 
        & LaBRI-IQDA~\cite{Commowick2021} & N/A & N/A & $51.51_{32.63}^{*}$ & N/A & N/A \\ 
        & Avg. of Experts~\cite{Commowick2021}  & N/A & N/A & $\mathbf{61.62_{37.11}}$ & N/A & N/A\\\midrule
        \multirow{5}{*}{\shortstack[l]{Task-\\ specific\\ SOTA}} & nnU-Net~\cite{Isensee2021} $\dagger$ & $75.81_{3.36}^{***}$ & $69.46_{11.44}^{***}$ & \multicolumn{1}{|c}{$54.15_{33.97}^{}$} & N/A & N/A \\
        &nnFormer~\cite{Zhou2023} $\dagger$ & $73.59_{5.04}^{***}$ & $68.13_{15.51}^{***}$ & \multicolumn{1}{|c}{$48.12_{33.53}^{}$} & N/A & N/A \\
        &UNETR~\cite{Hatamizadeh2022} $\dagger$ & $73.00_{4.92}^{***}$ & $70.12_{14.18}^{***}$ & \multicolumn{1}{|c}{$45.93_{39.24}^{**}$} & N/A & N/A \\
        &TransBTS~\cite{transbts} $\dagger$ & $73.94_{5.89}^{***}$ & $68.92_{14.77}^{***}$ & \multicolumn{1}{|c}{$46.10_{36.80}^{**}$} & N/A & N/A \\
        & TransUNet~\cite{Chen2021TransUNetTM} $\dagger$ & $72.90_{6.02}^{***}$ & $68.25_{14.90}^{***}$ & \multicolumn{1}{|c}{$42.47_{38.51}^{***}$} & N/A & N/A \\ \midrule
        \multirow{3}{*}{\shortstack[l]{Hetero-\\geneous\\methods}}&CoactSeg~\cite{Wu2023} $\ddagger$  & $74.51_{4.77}^{***}$ & $69.23_{14.26}^{***}$ & $47.23_{36.45}^{*}$ & $54.31_{27.91}^{***}$ & N/A \\
        &SegHeD\cite{basaran2024seghed} & $75.65_{4.74}$ & $76.14_{10.97}$ & $53.27_{34.70}$ & $59.20_{28.10}$ & $39.48_{28.41}$\\
        &\textbf{SegHeD+} & $\mathbf{79.35_{3.53}}$ & $\mathbf{76.82_{10.83}}$ & $55.02_{33.20}$ & $\mathbf{60.39_{28.10}}$ & $\mathbf{46.02_{21.50}}$\\ \bottomrule    
\end{tabular}
\end{table}

\subsubsection{All-lesion segmentation} 
The ``All" columns in Table~\ref{table:segresults} and Table~\ref{table:f1} demonstrate that SegHeD+ achieves significantly better performance in all-lesion segmentation compared to benchmark methods, in terms of the Dice score and $F_1$ score. On the longitudinal MS2015 dataset with all-lesion labels, SegHeD+ achieves Dice and $F_1$ scores of $76.39\%$ and $77.45\%$, respectively, presenting exceptional performance in lesion segmentation and detection, and significantly outperforming state-of the art models and CoactSeg. A component of SegHeD+ which improves longitudinal all-lesion segmentation performance is the inclusion of the previous timepoint segmentation as an input channel for the following timepoint prediction. This acts as a temporal guide for the network, and we see that predicted lesion volumes from SegHeD+ closely track the ground truth from the second timepoint onwards, explored in \ref{subsec:tempconsistency}.

\begin{figure}[!htp] \centering
\includegraphics[width=\linewidth]{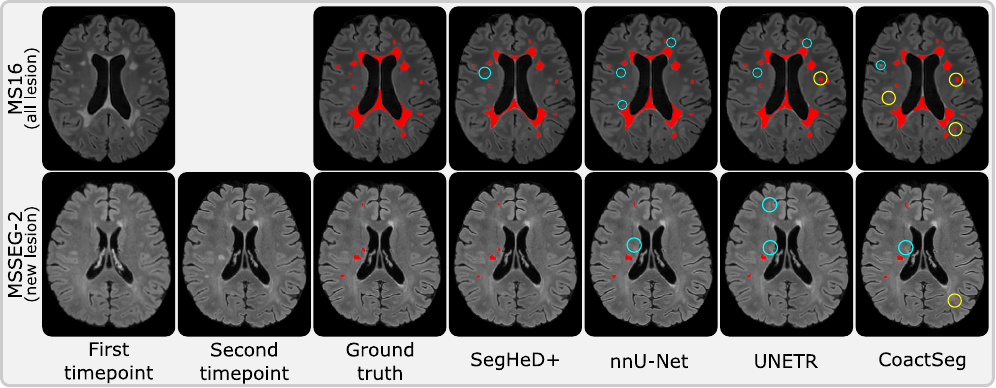}
\caption{Qualitative comparison of all-lesion (top row) and new-lesion (bottom row) segmentation performance. Yellow regions denote false positive segmentations, whereas cyan regions denote false negative segmentations. SegHeD+ exhibits fewer false segmentations. Best viewed online.} 
\label{fig:segmentations1}
\end{figure}

On the cross-sectional MS2016 test dataset, SegHeD+ achieves a Dice score of $85.18\%$, which is over $10\%$ higher than task-specific models and CoactSeg. Likewise, in lesion detection, SegHeD+ obtains an $F_1$ score of $76.82\%$, significantly outperforming SOTA. The improvements in the MS2015 and MS2016 dataset all-lesion segmentation performance can be attributed to various factors. Firstly, SegHeD+ enables the incorporation of heterogeneous data inputs, thus allowing for the inclusion of a larger number of images for model training. This allows for better model generalisation, as the model learns from more diverse images and from different scanners. We investigate the effect of this further in Section \ref{ablation:datasets} and Table~\ref{table:ablation_datasets}. Secondly, the integration of domain knowledge encoded via anatomical constraints contributes to the reduction of false positives and false negatives. Figure~\ref{fig:segmentations1} illustrates a qualitative comparison between SegHeD+ and selected benchmark methods, where we see a reduced amount of false positive and false negative segmentations.

\subsubsection{New-lesion segmentation}
The ``New" column in Table~\ref{table:segresults} demonstrates SegHeD+'s comparable performance to methods tailored for specific tasks and CoactSeg in the segmentation of new lesions. SegHeD+ achieves a Dice score of $50.52\%$ and an $F_1$ score of $55.02\%$, outperforming all SOTA and heterogeneous methods, and scoring slightly lower than the top methods tailored specifically for the new-lesion segmentation. Despite this SegHeD+ showcases the ability to handle multiple tasks using a single model.

\subsubsection{Vanishing-lesion segmentation}
The challenge of this task lies in the combined modeling of vanishing lesions along with all and new lesions. There are currently no established methods for segmenting disappearing lesions. As the last column of Table~\ref{table:segresults} shows, SegHeD+ achieves a Dice score of 43.84\% in this challenging task, showcasing an impressive improvement over SegHeD. This is largely attributed to the incorporation of LesionMix augmentation, providing additional vanishing lesions examples in the augmented datasets~\ref{table:lesionmix_dataset}. The asymmetry between the new and vanishing lesion segmentation Dice scores are due to two reasons. Firstly, new and all-lesion detection overlap in common features, making the task of new lesion segmentation easier. For new and all lesion segmentation, the network searches for hyperintense regions with respect to surrounding tissues. On the contrary, vanishing lesions manifest as normal tissue intensities, and require segmentation of hypo-intense tissue with respect to a previous timepoint. This contrast deems vanishing lesion segmentation to be more challenging than new lesion segmentation when combined with all lesion segmentation. Secondly, new-lesion detection is supported by two datasets (MSSEG-2, MSSEG-2+) in contrast to one vanishing dataset (VAN), thus utilising more new-lesion data. We believe that the results presented here provide useful insights for future work in dataset curation and benchmarking efforts.

\subsection{Ablation studies}
Ablation studies are conducted to examine various loss terms and the white matter mask input channel, LesionMix data augmentation, and effect of heterogeneous dataset inclusion.

\subsubsection{Effect of loss functions and white matter mask}
We provide the ablation results of the proposed loss functions in Table~\ref{table:ablation}. Each loss component plays a role in improving segmentation performance. Notably, new and vanishing-lesion segmentation performance increases when $\mathcal{L}_{Long}$ is introduced, all-lesion segmentation performance increases when $\mathcal{L}_{Vol}$ is introduced, and an overall increase occurs when $\mathcal{L}_{Spat}$ and $x_{wm}^{t2}$ are presented. In general, they lead to over 4\% increase in the Dice score for all-lesion segmentation on MS2015 and MS2016, and over 5\% improvement for new-lesion segmentation on the MSSEG-2 dataset.

\subsubsection{Effect of lesion-level data augmentation}
We also assess the effect of incorporating LesionMix as an offline data augmentation method. In the final row of Table~\ref{table:ablation}, we notice an overall increase in Dice score across all five datasets. Notably, LesionMix aids in segmenting vanishing lesions the most, boosting segmentation from 35\% to 43\% on the VAN dataset. This is expected, as LesionMix provides many augmentations with vanishing lesions across varying datasets, seen in Table~\ref{table:lesionmix_dataset}.

\begin{table}[htbp]
\begin{center}
\caption{Ablation study of proposed longitudinal, volumetric, and spatial loss functions ($\mathcal{L}_{Long}$, $\mathcal{L}_{Vol}$, $\mathcal{L}_{Spat}$), white matter mask input channel ($x_{wm}^{t2}$), and LesionMix augmentation.}
\setlength{\tabcolsep}{2mm}{
\centering
\begin{tabular}{cccccccccc} 
\toprule
\multicolumn{5}{c}{Ablation components} &\multicolumn{5}{c}{Dice score (\%)} \\
\cmidrule(r){1-5} \cmidrule(l){6-10}
\multirow{3}{*}{\vspace{0cm}\rotatebox[origin=l]{90}{\small  $\mathcal{L}_{Long}$}} & 
\multirow{3}{*}{\vspace{0cm}\rotatebox[origin=l]{90}{\small $\mathcal{L}_{Vol}$}} & 
\multirow{3}{*}{\vspace{0cm}\rotatebox[origin=l]{90}{\small $\mathcal{L}_{Spat}$}} & 
\multirow{3}{*}{\vspace{0cm}\rotatebox[origin=l]{90}{\small $x_{wm}^{t2}$}} & 
\multirow{3}{*}{\vspace{0cm}\rotatebox[origin=l]{90}{\small LesionMix}} & 
\multicolumn{5}{c}{Segmentation task (lesion type)} \\ 
\cmidrule(lr){6-10}
& & & & & \multicolumn{2}{c}{All} & \multicolumn{1}{c}{New} &\multicolumn{1}{c}{All\&New}  & \multicolumn{1}{c}{Vanishing} \\
\cmidrule(lr){6-7} \cmidrule(lr){8-8} \cmidrule(lr){9-9} \cmidrule(lr){10-10} 
& & & & & MS2015 & MS2016 & MSSEG-2 & MSSEG-2+ & VAN \\ 
\midrule

- & - & - & - & - & $72.98_{7.25}$ & $76.93_{9.01}$ & $45.28_{38.54}$ & $60.25_{20.04}$ & $30.97_{27.34}$\\
- & - & - & \checkmark & - & $73.35_{8.26}$ & $77.40_{9.05}$ & $46.15_{37.06}$ & $61.80_{20.90}$ & $31.50_{26.60}$\\
\checkmark & -  & - & - & - & $73.84_{6.99}$ & $77.38_{8.99}$ & $48.55_{34.80}$ & $64.86_{21.12}$ & $33.88_{28.97}$\\
\checkmark & \checkmark & - & - & - & $75.25_{7.83}$ & $81.99_{8.03}$ & $48.49_{35.00}$ & $65.01_{20.21}$ & $34.52_{30.63}$\\
\checkmark & \checkmark & \checkmark & - & - & $77.63_{6.62}$ & $82.71_{7.42}$& $48.56_{34.04}$ & $65.20_{19.79}$ & $34.90_{28.57}$\\
\checkmark & \checkmark & \checkmark & \checkmark & - & $77.95_{6.93}$ & $84.73_{7.12}$ &$48.64_{33.81}$ & $65.51_{19.07}$ & $35.23_{20.62}$\\
\checkmark & \checkmark & \checkmark & \checkmark & \checkmark & $78.57_{6.71}$ & $85.18_{7.10}$ &$50.52_{31.29}$ & $66.83_{18.98}$ & $43.84_{18.04}$\\
\bottomrule
\label{table:ablation}
\end{tabular}}
\end{center}
\end{table} 
\vspace{-1cm}
\subsubsection{Effect of integrating heterogeneous datasets}
\label{ablation:datasets}
We carry out an ablation study to analyse the effect of integrating heterogeneous datasets for lesion segmentation model training. This is performed with the proposed anatomical constraints and LesionMix augmentation. The impact on lesion segmentation when including varying amounts of heterogeneous data is presented in Table~\ref{table:ablation_datasets}. When using only the MS2015 and MS2016 datasets, we still see an improvement over task-specific SOTA methods due to the addition of anatomical constraints and additional augmentation. Most significantly, we notice an increase in performance when datasets with overlapping tasks are added. For example, when the MSSEG-2+ dataset is added, which includes the all and new-lesion segmentation tasks specifically, we see an improvement in all-lesion segmentation, as well as new-lesion segmentation.

\begin{table}[htbp]
\begin{center}
\caption{Ablation study showing improved performance when integrating heterogeneous datasets. All studies are implemented with anatomical constraints and LesionMix augmentation. N/A: output not available for the given dataset.}
\setlength{\tabcolsep}{1.4mm}{
\begin{tabular}{cccccccccc} 
\toprule
\multicolumn{5}{c}{Datasets included} &\multicolumn{5}{c}{Dice score (\%)} \\
\cmidrule(r){1-5} \cmidrule(l){6-10}
\multirow{3}{*}{\vspace{0cm}\rotatebox[origin=l]{90}{\small  MS2015}} & 
\multirow{3}{*}{\vspace{0cm}\rotatebox[origin=l]{90}{\small MS2016}} & 
\multirow{3}{*}{\vspace{0cm}\rotatebox[origin=l]{90}{\small MSSEG-2}} & 
\multirow{3}{*}{\vspace{0cm}\rotatebox[origin=l]{90}{\small MSSEG-2+}} & 
\multirow{3}{*}{\vspace{0cm}\rotatebox[origin=l]{90}{\small VAN}} & 
\multicolumn{5}{c}{Segmentation task (lesion type)} \\ 
\cmidrule(lr){6-10}
& & & & & \multicolumn{2}{c}{All} & \multicolumn{1}{c}{New} &\multicolumn{1}{c}{All\&New}  & \multicolumn{1}{c}{Vanishing} \\
\cmidrule(lr){6-7} \cmidrule(lr){8-8} \cmidrule(lr){9-9} \cmidrule(lr){10-10} 
& & & & & MS2015 & MS2016 & MSSEG-2 & MSSEG-2+ & VAN \\ 
\midrule
\checkmark & \checkmark & - & - & - & $76.81_{5.41}$ & $79.60_{7.18}$ & N/A & N/A & N/A \\
- & - & \checkmark & - & - & N/A & N/A & $49.82_{31.02}$ & N/A & N/A\\
\checkmark & \checkmark & \checkmark & - & - & $77.34_{6.19}$ & $79.98_{7.39}$ & $48.20_{32.58}$ & $62.18_{18.47}$ & N/A\\
\checkmark & \checkmark & \checkmark & \checkmark & - & $78.28_{6.58}$ & $85.26_{6.94}$ & $50.39_{30.58}$ & $67.21_{18.20}$ & N/A\\
\checkmark & \checkmark & \checkmark & \checkmark & \checkmark & $78.57_{6.71}$ & $85.18_{7.10}$ &$50.52_{31.29}$ & $66.83_{18.98}$ & $43.84_{18.04}$\\
\bottomrule
\label{table:ablation_datasets}
\end{tabular}}
\end{center}
\end{table}

\subsection{Improved temporal consistency}
\label{subsec:tempconsistency}
\begin{figure}[h!] \centering
\includegraphics[width=\linewidth]{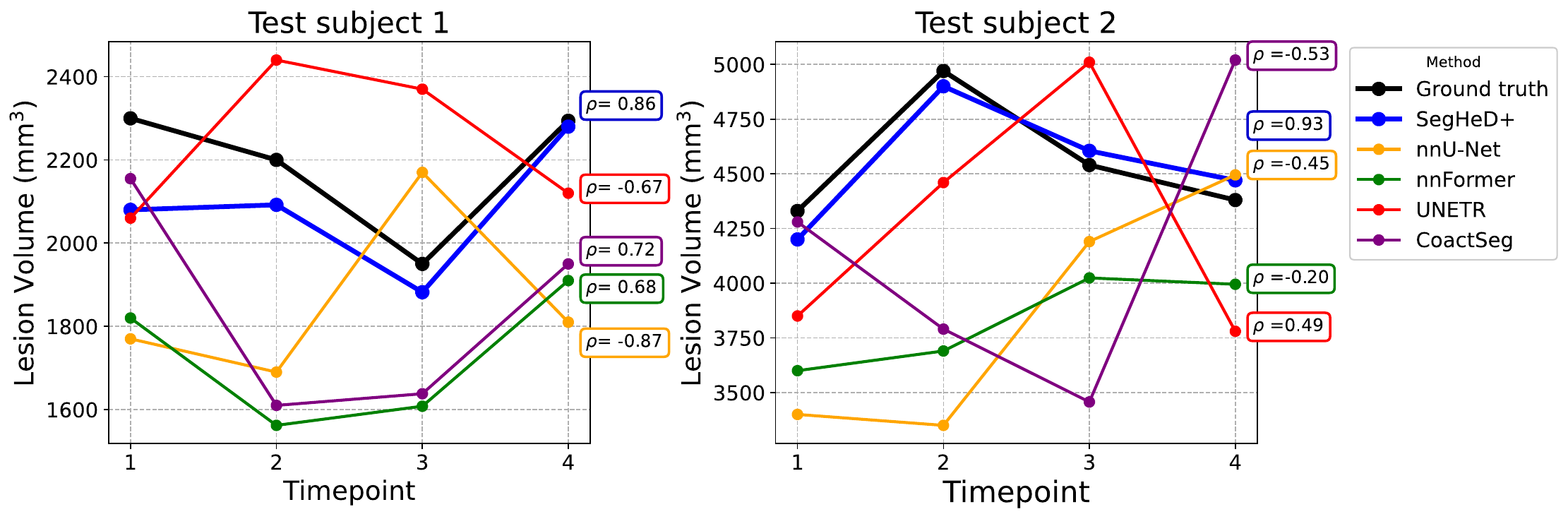}
\caption{Estimated lesion volumes across four time points for two test subjects. SegHeD+ (blue) results in predictions that are temporally more consistent with the ground truth (black), compared to competing methods. The $\rho$ value for each method indicates its Pearson's correlation coefficient with the ground truth, the higher the better.} 
\label{fig:consistency}
\end{figure}

We carry out a study to analyse the effect on the temporal consistency of lesion segmentation. Two subjects, each with four longitudinal scans, which were put aside for testing in the MS2015 dataset are utilised for this analysis.
Figure~\ref{fig:consistency} illustrates the predicted lesion volumes at various time points for the two individuals under examination. The figure shows that the lesion volumes estimated by SegHeD+ (depicted in blue) exhibit greater temporal coherence with the ground truth values (depicted in black) owing to the suggested anatomical constraints, resulting in the highest Pearson's correlation coefficient with the ground truth lesion volumes. When the all-lesion segmentations of previous timepoints are introduced in the inference stage, which occurs from the second timepoint onwards, SegHeD+ tracks the ground truth lesion volumes with less error. Enhancing temporal consistency aids in subsequent clinically relevant analyses for disease progression, such as assessing the annual rate of lesion shrinkage or expansion.

\section{Discussion}
\label{discussion}
We showcase that by integrating multiple heterogeneous datasets, adopting a multi-task learning approach, along with anatomical constraints and lesion-aware data augmentation, we can achieve promising results in multiple tasks simultaneously, including all, new and vanishing lesion segmentation tasks. While task-specific lesion segmentation models would typically focus on learning either hyperintense region features for segmenting new and all lesions, \emph{or} hypointense region features for segmenting vanishing lesions, SegHeD+ simultaneously learns both types of features, accounting for previous timepoints and surrounding tissues. SegHeD+ surpasses SOTA brain lesion segmentation models, which are trained in a task-specific manner, in terms of both lesion segmentation (Dice) and lesion detection ($F_1$) performance. 

There are a few limitations of this work. The first limitation is the relatively high computational cost during training. Utilising up to four 3D brain images as the model input leads to high model training times. A promising future avenue is to lower the computational burden associated with high-dimensional and high-resolution medical imaging data. Secondly, we provide a unique method for simulating a vanishing lesion dataset, VAN, through the use of a new lesion dataset, MSSEG-2. Vanishing and newly forming lesions show different lesion dynamics and characteristics~\cite{Molyneux1998}. A dedicated clinical dataset for vanishing lesions would enable more accurate modeling for real-life clinical use.

The case for a unified multi-task model which can outperform task-specific methods is not yet fully achieved. While SegHeD+ achieves the strongest performance in most of the tasks and datasets, some new-lesion methods still edge out SegHeD+, as shown in Table~\ref{table:segresults}. This performance gap may be reduced when more training datasets are provided for a specific task, as seen in Table~\ref{table:ablation_datasets}. Alongside new and vanishing lesions, growing and shrinking lesion detection is of paramount importance for analysing the progression of MS. However, the development of state-of-the-art machine learning methods for clinical applications depend heavily on data availability, and currently there are no public datasets for analysing growing and/or shrinking MS lesions.

\section{Conclusion}
\label{conclusion}
In this work we introduce SegHeD+, an innovative multi-task approach for segmenting MS lesions that can effectively learn from diverse and heterogeneous data. SegHeD+ can accurately segment all types of lesions, including new and disappearing ones, in both cross-sectional and longitudinal datasets. Evaluations on five MS lesion datasets demonstrate that SegHeD+ achieves superior performance compared to other segmentation techniques for segmenting all types of lesions, and shows comparable results for identifying new lesions. Considering the current lack of large MS lesion datasets, SegHeD+'s ability to utilise diverse data sources will significantly enhance current MS imaging research and simplify the analysis of large-scale multi-site datasets with varying data structures and annotations.


\subsubsection{Acknowledgements}  This work is supported by the UKRI CDT in AI for Healthcare \href{http://ai4health.io}{http://ai4health.io} (Grant No. EP/S023283/1). W. Bai is co-funded by EPSRC DeepGeM Grant (EP/W01842X/1) and NIHR Imperial Biomedical Research Centre (BRC). The views expressed are those of the authors and not necessarily those of the NIHR or the Department of Health and Social Care. 

%
%
\bibliographystyle{ieeetr}
\bibliography{refs}

\begin{thebibliography}{10}

\bibitem{Lassmann2018}
H.~Lassmann, ``Multiple sclerosis pathology,'' {\em Cold Spring Harbor Perspectives in Medicine}, vol.~8, no.~3, 2018.

\bibitem{Carass2015}
A.~Carass, S.~Roy, A.~Jog, {\em et~al.}, ``{Longitudinal multiple sclerosis lesion segmentation: Resource and challenge},'' {\em NeuroImage}, 2017.

\bibitem{Commowick2018}
O.~Commowick, A.~Istace, M.~Kain, {\em et~al.}, ``Objective evaluation of multiple sclerosis lesion segmentation using a data management and processing infrastructure,'' {\em Scientific Reports}, 2018.

\bibitem{Commowick2021}
O.~Commowick, A.~Masson, B.~Combes, {\em et~al.}, ``{MICCAI 2021 MSSEG-2 challenge quantitative results},'' 2021.

\bibitem{Eckert2018}
S.~Eckert, B.~Weinstock-Guttman, C.~Kolb, and D.~Hojnacki, ``Disappearing brainstem mri lesions in multiple sclerosis (p3.353),'' {\em Neurology}, vol.~90, no.~15\_supplement, p.~P3.353, 2018.

\bibitem{Dwyer2018}
M.~G. Dwyer, N.~Bergsland, D.~P. Ramasamy, D.~Jakimovski, B.~Weinstock-Guttman, and R.~Zivadinov, ``Atrophied brain lesion volume: A new imaging biomarker in multiple sclerosis,'' {\em Journal of Neuroimaging}, vol.~28, no.~5, pp.~490--495, 2018.

\bibitem{prineas199}
J.~W. Prineas, R.~O. Barnard, T.~Revesz, {\em et~al.}, ``Multiple sclerosis: pathology of recurrent lesions,'' {\em Brain}, vol.~116, no.~3, 1993.

\bibitem{kamnitsas2017}
K.~Kamnitsas, C.~Ledig, V.~F. Newcombe, {\em et~al.}, ``{Efficient multi-scale 3D CNN with fully connected CRF for accurate brain lesion segmentation},'' {\em Medical Image Analysis}, vol.~36, 2017.

\bibitem{basaran2023}
B.~D. Basaran, W.~Zhang, M.~Qiao, {\em et~al.}, ``{LesionMix: A lesion-level data augmentation method for medical image segmentation},'' in {\em MICCAI DALI Workshop}, 2023.

\bibitem{Elliott2013}
C.~Elliott, D.~L. Arnold, D.~L. Collins, and T.~Arbel, ``{Temporally consistent probabilistic detection of new multiple sclerosis lesions in brain MRI},'' {\em IEEE Transactions on Medical Imaging}, vol.~32, no.~8, 2013.

\bibitem{Jain2016}
S.~Jain, A.~Ribbens, D.~M. Sima, {\em et~al.}, ``{Two time point MS lesion segmentation in brain MRI: An expectation-maximization framework},'' {\em Frontiers in Neuroscience}, vol.~10, 2016.

\bibitem{Denner2021}
S.~Denner, A.~Khakzar, M.~Sajid, {\em et~al.}, ``Spatio-temporal learning from longitudinal data for multiple sclerosis lesion segmentation,'' in {\em Brainlesion: Glioma, Multiple Sclerosis, Stroke and Traumatic Brain Injuries} (A.~Crimi and S.~Bakas, eds.), 2021.

\bibitem{Isensee2021}
F.~Isensee, P.~F. Jaeger, S.~A.~A. Kohl, {\em et~al.}, ``{nnU-Net: a self-configuring method for deep learning-based biomedical image segmentation},'' {\em Nature Methods}, 2021.

\bibitem{Basaran2022}
B.~D. Basaran, P.~M. Matthews, and W.~Bai, ``New lesion segmentation for multiple sclerosis brain images with imaging and lesion-aware augmentation,'' {\em Frontiers in Neuroscience}, vol.~16, 2022.

\bibitem{Wu2023}
Y.~Wu, Z.~Wu, H.~Shi, {\em et~al.}, ``{CoactSeg: Learning from heterogeneous data for new multiple sclerosis lesion segmentation},'' in {\em Medical Image Computing and Computer Assisted Intervention}, 2023.

\bibitem{Liu2023}
J.~Liu, Y.~Zhang, J.~Chen, {\em et~al.}, ``{CLIP-driven universal model for organ segmentation and tumor detection},'' {\em International Conference on Computer Vision}, 2023.

\bibitem{Shi2021}
G.~Shi, L.~Xiao, Y.~Chen, and S.~K. Zhou, ``Marginal loss and exclusion loss for partially supervised multi-organ segmentation,'' {\em Medical Image Analysis}, vol.~70, 2021.

\bibitem{Butoi_2023_ICCV}
V.~I. Butoi, J.~J.~G. Ortiz, T.~Ma, M.~R. Sabuncu, J.~Guttag, and A.~V. Dalca, ``Universeg: Universal medical image segmentation,'' in {\em International Conference on Computer Vision}, 2023.

\bibitem{lin2023samus}
X.~Lin, Y.~Xiang, L.~Zhang, X.~Yang, Z.~Yan, and L.~Yu, ``Samus: Adapting segment anything model for clinically-friendly and generalizable ultrasound image segmentation,'' {\em arXiv: 2309.06824}, 2023.

\bibitem{Dalca2018}
A.~V. Dalca, J.~Guttag, and M.~R. Sabuncu, ``Anatomical priors in convolutional networks for unsupervised biomedical segmentation,'' in {\em Computer Vision and Pattern Recognition}, 2018.

\bibitem{Strumia2016}
M.~Strumia, F.~R. Schmidt, C.~Anastasopoulos, {\em et~al.}, ``{White matter MS-lesion segmentation using a geometric brain model},'' {\em IEEE Transactions on Medical Imaging}, vol.~35, no.~7, 2016.

\bibitem{Hirsch2021}
L.~Hirsch, Y.~Huang, and L.~C. Parra, ``{Segmentation of MRI head anatomy using deep volumetric networks and multiple spatial priors},'' {\em Journal of Medical Imaging}, vol.~8, no.~3, 2021.

\bibitem{Zhang2021}
X.~Zhang, C.~Liu, N.~Ou, {\em et~al.}, ``{CarveMix: A simple data augmentation method for brain lesion segmentation},'' in {\em Medical Image Computing and Computer Assisted Intervention}, 2021.

\bibitem{Reinhold2021}
J.~C. Reinhold, A.~Carass, and J.~L. Prince, ``{A structural causal model for MR images of multiple sclerosis},'' in {\em Medical Image Computing and Computer Assisted Intervention}, 2021.

\bibitem{Basaran2022pseudo}
B.~D. Basaran, M.~Qiao, P.~M. Matthews, and W.~Bai, ``Subject-specific lesion generation and pseudo-healthy synthesis for multiple sclerosis brain images,'' in {\em Simulation and Synthesis in Medical Imaging}, 2022.

\bibitem{basaran2024seghed}
B.~D. Basaran, X.~Zhang, P.~M. Matthews, and W.~Bai, ``{SegHeD}: Segmentation of heterogeneous data for multiple sclerosis lesions with anatomical constraints,'' {\em arXiv preprint arXiv:2410.01766}, 2024.

\bibitem{Billot2023}
B.~Billot, D.~N. Greve, O.~Puonti, {\em et~al.}, ``{SynthSeg: Segmentation of brain MRI scans of any contrast and resolution without retraining},'' {\em Medical Image Analysis}, vol.~86, 2023.

\bibitem{Milletari2016}
F.~Milletari, N.~Navab, and S.-A. Ahmadi, ``{V-net: Fully convolutional neural networks for volumetric medical image segmentation},'' in {\em International Conference on 3D Vision}, 2016.

\bibitem{Homssi2023}
M.~Homssi, E.~Sweeney, E.~Demmon, {\em et~al.}, ``{Evaluation of the statistical detection of change algorithm for screening patients with MS with new lesion activity on longitudinal brain MRI},'' {\em American Journal of Neuroradiology}, vol.~44, no.~6, 2023.

\bibitem{Filippi1995}
M.~Filippi, M.~A. Horsfield, P.~S. Tofts, {\em et~al.}, ``{Quantitative assessment of MRI lesion load in monitoring the evolution of multiple sclerosis},'' {\em Brain}, vol.~118, no.~6, 1995.

\bibitem{Filippi2010}
M.~Filippi and F.~Agosta, ``Imaging biomarkers in multiple sclerosis,'' {\em Journal of Magnetic Resonance Imaging}, vol.~31, no.~4, pp.~770--788, 2010.

\bibitem{Tiu2022}
V.~E. Tiu, I.~Enache, C.~A. Panea, C.~Tiu, and B.~O. Popescu, ``Predictive mri biomarkers in ms—a critical review,'' {\em Medicina}, vol.~58, no.~3, 2022.

\bibitem{Sethi2017}
V.~Sethi, G.~Nair, M.~Absinta, {\em et~al.}, ``Slowly eroding lesions in multiple sclerosis,'' {\em Multiple Sclerosis Journal}, vol.~23, no.~3, 2017.

\bibitem{Genovese2019}
A.~V. Genovese, J.~Hagemeier, N.~Bergsland, {\em et~al.}, ``{Atrophied brain T2 lesion volume at MRI is associated with disability progression and conversion to secondary progressive multiple sclerosis},'' {\em Radiology}, vol.~293, no.~2, 2019.

\bibitem{Molyneux1998}
P.~D. Molyneux, M.~Filippi, F.~Barkhof, {\em et~al.}, ``Correlations between monthly enhanced mri lesion rate and changes in t2 lesion volume in multiple sclerosis,'' {\em Annals of Neurology}, vol.~43, no.~3, 1998.

\bibitem{Bengio2009}
Y.~Bengio, J.~Louradour, R.~Collobert, and J.~Weston, ``Curriculum learning,'' in {\em International Conference on Machine Learning}, 2009.

\bibitem{Telea2004}
A.~Telea, ``An image inpainting technique based on the fast marching method,'' {\em Journal of Graphics Tools}, 2004.

\bibitem{Coupe2008}
P.~Coupe, P.~Yger, S.~Prima, P.~Hellier, C.~Kervrann, and C.~Barillot, ``An optimized blockwise nonlocal means denoising filter for 3-d magnetic resonance images,'' {\em IEEE Transactions on Medical Imaging}, vol.~27, no.~4, 2008.

\bibitem{Commowick2012}
O.~Commowick, N.~Wiest-Daesslé, and S.~Prima, ``Block-matching strategies for rigid registration of multimodal medical images,'' in {\em IEEE International Symposium on Biomedical Imaging}, pp.~700--703, 2012.

\bibitem{Manjon2016}
J.~V. Manjón and P.~Coupé, ``volbrain: An online mri brain volumetry system,'' {\em Frontiers in Neuroinformatics}, vol.~10, 2016.

\bibitem{Tustison2010}
N.~J. Tustison, B.~B. Avants, P.~A. Cook, Y.~Zheng, A.~Egan, P.~A. Yushkevich, and J.~C. Gee, ``N4itk: Improved n3 bias correction,'' {\em IEEE Transactions on Medical Imaging}, vol.~29, no.~6, 2010.

\bibitem{Doshi2013}
J.~Doshi, G.~Erus, Y.~Ou, B.~Gaonkar, and C.~Davatzikos, ``{Multi-Atlas Skull-Stripping},'' {\em Academic Radiology}, vol.~20, no.~12, pp.~1566--1576, 2013.

\bibitem{ITKSNAP}
P.~A. Yushkevich, Y.~Gao, and G.~Gerig, ``Itk-snap: An interactive tool for semi-automatic segmentation of multi-modality biomedical images,'' in {\em International Conference of the IEEE Engineering in Medicine and Biology Society}, 2016.

\bibitem{Fonov2009}
V.~Fonov, A.~Evans, R.~McKinstry, {\em et~al.}, ``Unbiased nonlinear average age-appropriate brain templates from birth to adulthood,'' {\em NeuroImage}, 2009.

\bibitem{Kamraoui2022}
R.~A. Kamraoui, B.~Mansencal, J.~V. Manjon, and P.~Coupé, ``Longitudinal detection of new ms lesions using deep learning,'' {\em Frontiers in Neuroimaging}, vol.~1, 2022.

\bibitem{Zhou2023}
H.-Y. Zhou, J.~Guo, Y.~Zhang, {\em et~al.}, ``{nnFormer: Volumetric medical image segmentation via a 3D transformer},'' {\em IEEE Transactions on Image Processing}, 2023.

\bibitem{Hatamizadeh2022}
A.~Hatamizadeh, Y.~Tang, V.~Nath, {\em et~al.}, ``{UNETR: Transformers for 3D medical image segmentation},'' in {\em Proceedings of the IEEE/CVF Winter Conference on Applications of Computer Vision}, 2022.

\bibitem{zhang2021msseg}
H.~Zhang, H.~Li, and I.~Oguz, ``{Segmentation of new MS lesions with tiramisu and 2.5D stacked slices},'' {\em MSSEG-2 challenge proceedings: Multiple sclerosis new lesions segmentation challenge using a data management and processing infrastructure}, vol.~61, 2021.

\bibitem{transbts}
W.~Wang, C.~Chen, M.~Ding, H.~Yu, S.~Zha, and J.~Li, ``{TransBTS: Multimodal brain tumor segmentation using transformer},'' in {\em Medical Image Computing and Computer-Assisted Intervention}, 2021.

\bibitem{Chen2021TransUNetTM}
J.~Chen, Y.~Lu, Q.~Yu, X.~Luo, E.~Adeli, {\em et~al.}, ``Transunet: Transformers make strong encoders for medical image segmentation,'' in {\em ICML Interpretable Machine Learning in Healthcare Workshop}, 2021.

\end{thebibliography}

\end{document}